\documentclass[preprint,10pt,authoryear]{elsarticle}

\usepackage{graphicx}%
\usepackage{dblfloatfix}% improve placement of two-column floats (figure*)
\usepackage{multirow}%
\usepackage{amsmath,amssymb,amsfonts}%
\usepackage{mathrsfs}%
\usepackage{dsfont}% \mathds{1}: indicator function
\usepackage{xcolor}%
\usepackage{textcomp}%
\usepackage{booktabs}%
\usepackage{algorithm}%
\usepackage{algorithmicx}%
\usepackage{algpseudocode}%
\usepackage{placeins}%
\usepackage{adjustbox}% shrinks a table only when it is wider than the text block
\usepackage[hidelinks]{hyperref}%
\makeatletter% vector fonts for elsarticle's T1 encoding, unless the 'times' option is used
\@ifpackageloaded{times}{}{\@ifpackageloaded{txfonts}{}{\usepackage{lmodern}}}
\makeatother
\makeatletter% first-page footer: page number only (no journal name)
\def\ps@pprintTitle{\let\@oddhead\@empty\let\@evenhead\@empty\def\@oddfoot{\hfil\thepage\hfil}\let\@evenfoot\@oddfoot}
\makeatother

\begin{document}

\begin{frontmatter}

\title{Compression of Vision Foundation Models for Retinal OCT via Growing-Neural-Gas Codebooks and Graph-Guided Masked Image Modeling}

\author[1]{Ali Yavari\corref{cor1}}
\ead{ali.yavari@meduniwien.ac.at}
\cortext[cor1]{Corresponding author.}
\author[2]{Tilman Schmoll}
\author[1]{Rainer A. Leitgeb}
\author[1]{Wolfgang Drexler}
\author[1]{Thomas Schlegl}

\affiliation[1]{organization={Center for Medical Physics and Biomedical Engineering, Medical University of Vienna},
            addressline={W{\"a}hringer G{\"u}rtel 18-20},
            city={Vienna},
            postcode={1090},
            country={Austria}}

\affiliation[2]{organization={Carl Zeiss Meditec AG},
            addressline={Rudolf-Eber-Stra{\ss}e 11},
            city={Oberkochen},
            postcode={73447},
            country={Germany}}

\begin{abstract}
Vision foundation models for retinal imaging are large and computationally demanding, limiting point-of-care deployment. We ask how much of such a model's downstream utility survives compression into a smaller network. We propose a topology-aware discretization framework that distills a continuous foundation model into a discrete codebook, then uses it as the reconstruction target for masked image modeling (MIM). It combines gradient-free codebook learning via Growing Neural Gas, which yields an adjacency graph over the codes, with a graph-distance soft cross-entropy loss that replaces one-hot MIM targets with topologically graded supervision. With student pretraining on 383 public Duke iAMD optical coherence tomography (OCT) volumes, the pipeline distills an 86M-parameter teacher into a 22M-parameter student with $\geq$3.6$\times$ fewer FLOPs. The student performs close to, and in some settings above, its teacher across three classification and two segmentation benchmarks, with the magnitude and direction of the difference depending on the teacher, task, metric, and student capacity. On the external OLIVES benchmark, it surpasses its MIRAGE teacher on every metric (98.06\% vs.\ 95.22\% balanced accuracy). It works with two Vision Transformer (ViT-Base) teachers, retinal (MIRAGE) and natural-image (DINOv3); with the out-of-domain teacher, the student outperforms it on retinal-layer segmentation, indicating that the added value of retina-specific pretraining is task-dependent: largest for segmentation, while a general-domain teacher is competitive for classification. These results yield compact retinal models potentially suitable for deployment, transferring across OCT datasets, devices, and clinical tasks, including three datasets not used in the student's pretraining.
\end{abstract}


\end{frontmatter}

\section{Introduction}\label{sec1}

Retinal diseases including age-related macular degeneration (AMD), diabetic macular edema (DME), and choroidal neovascularization (CNV) impose a major and growing clinical burden worldwide, driven by population aging and the diabetes epidemic, and represent leading causes of vision impairment and blindness \citep{Wong2014GlobalPrevalenceAMD}. Optical coherence tomography (OCT) has become the indispensable workhorse imaging modality for screening, diagnosis, and longitudinal monitoring of these conditions, delivering high-resolution, non-invasive cross-sectional visualization of retinal microstructure that directly informs anti-VEGF treatment decisions and disease activity assessment \citep{flaxel2020amdppp}. Deep learning has become integral to automated OCT interpretation, yet conventional supervised approaches remain constrained by the scarcity of large, expert-annotated retinal datasets, motivating a shift toward self-supervised pretraining that learns directly from unlabeled scans \citep{morano2025mirage}. Within this paradigm, Vision Transformer-based foundation models—including the general-domain DINOv3 \citep{simeoni2025dinov3} and retina-specific models such as RETFound \citep{zhou2023foundation}, VisionFM \citep{qiu2024development}, EyeCLIP \citep{shi2025eyeclip}, and MIRAGE \citep{morano2025mirage}—have demonstrated that large-scale self-supervised pretraining yields transferable representations that consistently outperform models trained from scratch or pretrained on ImageNet across diagnostic and prognostic retinal tasks. A single such backbone can deliver strong performance across both classification and segmentation of OCT images, making vision foundation models an increasingly attractive basis for retinal image analysis.

Despite this promise, the foundation models driving these gains are typically large, GPU-intensive architectures, and their substantial computational and infrastructure requirements remain a major obstacle to real-time deployment on portable devices or in resource-constrained clinical settings \citep{jin2025systematic, chia2024foundation}. Compounding this barrier, building a foundation model with broad generalizability typically demands pretraining on large, heterogeneous datasets pooled across institutions and imaging devices, a scale of data acquisition and curation that is logistically and financially prohibitive for most individual research groups or clinical centers \citep{chia2024foundation, Zhou2026PretrainingDataEffectsRetinalFoundationModels}.
A natural avenue to sidestep the data constraint is masked image modeling (MIM), a self-supervised paradigm that eliminates the need for labeled data and, in principle, allows pretraining on more modest unlabeled datasets. Pixel-level MIM approaches reconstruct raw image patches from masked inputs, an objective that is easy to optimize but biases the encoder toward low-level texture statistics rather than high-level semantic features \citep{He2022Masked}. Lifting the reconstruction target to the semantic level — by predicting discrete visual tokens produced by a vector-quantized tokenizer rather than raw pixels — was proposed as a remedy \citep{Bao2022BEiT, Ramesh2021DALLE, Peng2023BEiTv2}; yet standard vector quantization imposes no relational structure among learned tokens, and the one-hot MIM target treats every incorrect code as equally wrong. Crucially, both MIM generations were developed for large, GPU-intensive encoders, and neither was designed to transfer the semantic knowledge embedded in those models into a lightweight backbone suitable for point-of-care inference. Distillation into compact Vision Transformers has been studied on natural images, for example by DeiT, which transfers a teacher's class predictions through a dedicated distillation token during supervised ImageNet training \citep{touvron2021deit}, and by TinyMIM, which transfers token relations from larger MIM-pretrained ViTs to smaller ones on ImageNet \citep{ren2023tinymim}. Both were developed on large natural-image corpora with labels or continuous targets; they do not consider distillation from a foundation model through graph-structured discrete targets on a small, unlabeled in-domain medical cohort.

To address this gap, we propose a topology-aware discrete distillation framework that compresses the knowledge of a continuous vision foundation model into a compact discrete codebook, which then serves as the reconstruction target for masked image modeling. Rather than relearning representations from scratch, a student encoder is pretrained to predict the codes of a frozen teacher, transferring the teacher's semantic knowledge into a more compact backbone, as in knowledge distillation \citep{phuong2019towards, tian2020contrastive}; at its smallest, this yields a model with a reduced parameter footprint potentially suitable for deployment. The framework rests on three components. First, we learn the codebook with Growing Neural Gas (GNG), a gradient-free vector quantization method: the codebook grows adaptively through topology-driven insertion and pruning. Second, GNG yields not only the codes but also an adjacency graph linking them, endowing the discrete vocabulary with relational structure that standard tokenizers lack. Third, we exploit this structure through a graph-distance soft cross-entropy loss that replaces the flat one-hot MIM target with topologically graded supervision, so that predicting a code near the correct one on the graph is penalized less than predicting a distant one—turning masked prediction into structured prediction over the learned codebook. We instantiate the framework with two teachers from different domains—a retina-specific foundation model (MIRAGE; \citealp{morano2025mirage}) and a general-domain natural-image model (DINOv3; \citealp{simeoni2025dinov3})—and pretrain the student using only a single small public OCT dataset, Duke iAMD \citep{Farsiu2014AMD}. Both teachers evaluated here are ViT-Base with $D{=}768$, and the same pipeline applies unchanged to each.

We first evaluate the framework along two axes: how much it compresses, and how much of the teacher's downstream utility the compressed student preserves. Using this pipeline, an 86M-parameter teacher is distilled into a 22M-parameter student—roughly a fourfold reduction in size—which we assess across three OCT classification benchmarks and two segmentation benchmarks, spanning multiple datasets, imaging devices, and clinical tasks. Across these benchmarks the compact student achieved performance close to, and on individual tasks above, the teacher's, although the magnitude and direction of the difference depended on the teacher, task, metric, and student capacity. The same result holds for both the retina-specific and the natural-image teacher, which is consistent with the gains arising from the distillation framework rather than from any single backbone. Because the same pipeline runs unchanged with both teachers, it also lets us ask a second question: for which tasks a retina-specific foundation model still adds value relative to a general-domain backbone passed through the same pipeline. Taken together, these findings offer a route to compact retinal image models potentially suitable for deployment and, more broadly, a recipe for compressing medical vision foundation models into efficient backbones without re-collecting the large, multi-source corpora their original pretraining required.

\section{Materials and methods}\label{sec4}

\subsection{Framework overview}\label{subsec:overview}
\begin{figure*}[!htbp]
    \centering
    \includegraphics[width=\linewidth,height=0.58\textheight,keepaspectratio]{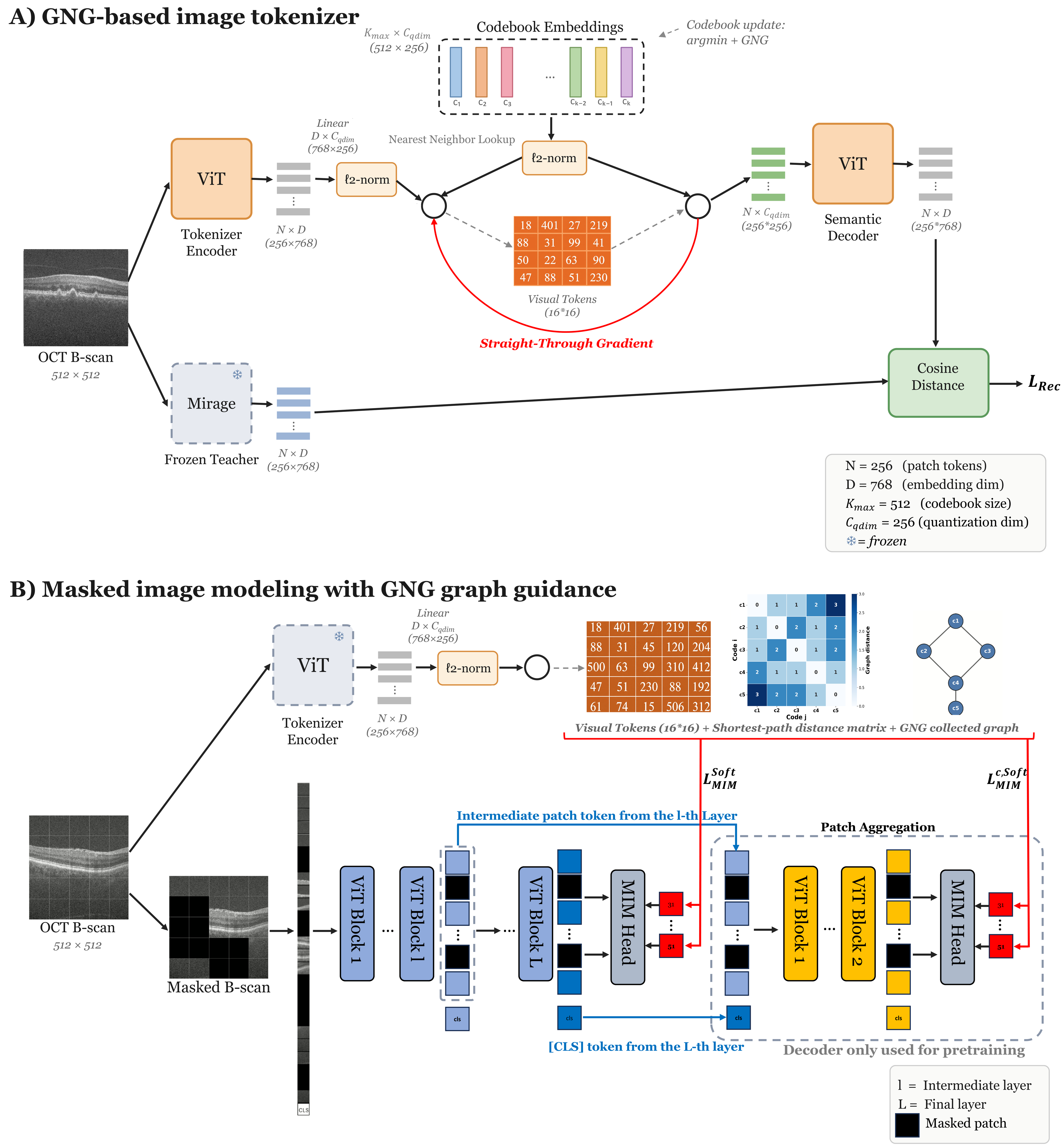}
    \caption{
    \textbf{Overview of the GNG-guided pretraining framework.}
    {\bfseries A} GNG-based image tokenizer: a ViT encoder's patch embeddings are quantized against a codebook grown by Growing Neural Gas (GNG) to $K_{\max}{=}512$ codes in a $C_{\mathrm{qdim}}{=}256$-dimensional space (straight-through gradient), and a decoder reconstructs the frozen teacher's embeddings under a cosine loss $\mathcal{L}_{\mathrm{rec}}$; GNG also induces the adjacency graph used in \textbf{B}.
    {\bfseries B} Masked image modeling with GNG graph guidance: masked-token targets are softened by the graph distance between codes, giving a patch-level objective $\mathcal{L}^{\mathrm{soft}}_{\mathrm{MIM}}$ and, via class-token aggregation, a global objective $\mathcal{L}^{c,\mathrm{soft}}_{\mathrm{MIM}}$. Only the student encoder is kept for downstream tasks.}
    \label{fig:framework}
\end{figure*}

We build on a two-stage self-supervised framework for learning transferable representations from retinal OCT B-scans (Fig.~\ref{fig:framework}). In the first stage, an image tokenizer distills the representation space of a frozen vision foundation model (the teacher) into a discrete codebook learned with Growing Neural Gas (GNG; \citealp{Fritzke1995}), which also induces an adjacency graph over the codes (Section~\ref{subsec:image_tokenizer}). In the second stage, the resulting tokens serve as prediction targets for masked image modeling (MIM), through which a more compact backbone (the student) is pretrained on unlabeled OCT data with a graph-distance soft cross-entropy objective (Section~\ref{subsec:mim_models}). The pretrained student is then evaluated on image-level classification and pixel-level segmentation.

\subsection{Pretraining dataset}\label{subsec:pretraining_data}

All self-supervised pretraining experiments were conducted exclusively on 
the Duke intermediate age-related macular degeneration dataset (Duke iAMD) 
\citep{Farsiu2014AMD}, a publicly available SD-OCT cohort collected across 
four clinical sites in the United States. The dataset encompasses 384 
volumetric SD-OCT scans from 384 subjects, comprising 269 patients 
diagnosed with intermediate AMD and 115 age-matched healthy controls. All 
volumes were acquired using a Bioptigen SD-OCT system (Bioptigen Inc., 
Research Triangle Park, NC, USA). Each volume covers a retinal field of 
view of approximately $6.7 \times 6.7$~mm\textsuperscript{2} centered on
the fovea and consists of 100 B-scans, each of $512 \times 1000$ pixels 
(axial $\times$ lateral).

In addition to subject-level diagnostic labels, the dataset provides 
dense, expert-annotated boundary delineations defining three anatomically 
distinct retinal regions: (i) the layer spanning from the inner limiting 
membrane (ILM) to the inner boundary of the retinal pigment epithelium
and drusen complex (RPEDC), (ii) the RPEDC itself, bounded inferiorly by 
the outer surface of Bruch's membrane (BM), and (iii) the sub-BM region. 
These boundary annotations were used to define the segmentation targets 
in all downstream evaluation experiments. One volume was excluded prior to any model training owing to a detected 
annotation inconsistency, yielding a final pretraining corpus of 383 
volumes corresponding to approximately 38,300 B-scans. 

Because no labels are used during pretraining, reusing Duke iAMD as a downstream benchmark introduces no label leakage. All 383 volumes were therefore available during self-supervised pretraining before the downstream train/validation/test split was formed, making Duke iAMD a label-free transductive evaluation. This follows the established masked-image-modeling protocol used by BEiT-2: its VQ-KD tokenizer and MIM encoder are pretrained on ImageNet-1K and the representation is subsequently fine-tuned or linearly probed on ImageNet-1K \citep{Peng2023BEiTv2}.

\subsection{Benchmark datasets}\label{subsec:benchmark_datasets}
We evaluated the pretrained backbone on five benchmarks drawn from four publicly available OCT datasets, spanning two task families: three for image-level classification (Kermany, \citealp{kermany2018labeled, kermany2018identifying}; OLIVES, \citealp{prabhushankar2022olives}; Duke iAMD, \citealp{Farsiu2014AMD}) and two for pixel-level segmentation (AROI, \citealp{Melinscak2021AROI}; Duke iAMD, \citealp{Farsiu2014AMD}), summarized in Table~\ref{tab:datasets}. Together they cover multiple retinal pathologies, three distinct acquisition devices, and independent acquisition sites. Duke iAMD is evaluated transductively (Section~\ref{subsec:pretraining_data}), whereas Kermany, OLIVES, and AROI, none of which was used for self-supervised pretraining, provide cross-dataset transfer evaluations. For the volumetric classification datasets (Duke iAMD and OLIVES) we restricted the analysis to the central B-scan of each volume, consistent with prior work \citep{morano2025mirage, zhou2023foundation}. For the segmentation datasets, all annotated B-scans were used, with each patient's B-scans assigned to a single partition, in line with standard practice in which retinal layer and lesion segmentation are performed B-scan-wise rather than at the volume level \citep{morano2025mirage, he2021structured, fazekas2023segmentation, roy2017relaynet}. Where official train/validation/test splits were available we adopted them directly. For Duke iAMD and the segmentation datasets we used a patient-level stratified split; for OLIVES we used the label-stratified patient-level split described in Section~\ref{subsec:implementation_details}.

\begin{table*}[!t]
\centering
\caption{Benchmark datasets used for downstream evaluation. Scan counts are B-scans; volume/subject counts are given in parentheses where applicable. Segmentation used only the expert-annotated B-scans: $1{,}105$ for AROI and $29{,}291$ for Duke iAMD ($74$--$77$ per volume).}
\label{tab:datasets}
\begin{adjustbox}{max width=\linewidth}
\begin{tabular}{@{}llrp{5.1cm}@{}}
\toprule
Dataset & Device & Scans (n) & Classes / Labels (n) \\
\midrule
\multicolumn{4}{l}{\textit{Classification}}\\
Kermany   & Spectralis        & 84{,}484           & CNV (37{,}455), DME (11{,}598), Drusen (8{,}866), Normal (26{,}565) \\
OLIVES    & Spectralis        & (1{,}594 vols.)     & DME (934), DR (660) \\
Duke iAMD & Bioptigen SD-OCT  & (383 vols.)         & Control (114), iAMD (269) \\
\midrule
\multicolumn{4}{l}{\textit{Segmentation}}\\
AROI      & Zeiss Cirrus      & 2{,}944 (23 pts.)   & Background (above ILM), ILM--IPL/INL, IPL/INL--RPE, RPE--BM, sub-BM, PED, SRF, IRF \\
Duke iAMD & Bioptigen SD-OCT  & 38{,}300 (383 vols.) & Background, ILM--RPEDC, RPEDC--BM, sub-BM \\
\bottomrule
\end{tabular}
\end{adjustbox}
\end{table*}

\subsection{GNG-based image tokenizer}\label{subsec:image_tokenizer}
The first stage learns a discrete image tokenizer that distills a frozen
vision foundation model into a codebook. We describe the tokenizer for
a generic frozen teacher $T$ and instantiate $T$ with two backbones: MIRAGE
\citep{morano2025mirage}, which is, to our knowledge, the most recent
published retinal vision foundation model, and the general-domain DINOv3
\citep{simeoni2025dinov3}. Each teacher is distilled into its own,
independently trained tokenizer, which comprises three modules
(Fig.~\ref{fig:framework}-A). (i) A tokenizer encoder $\mathrm{Enc}_\phi$ (with parameters $\phi$) embeds an input B-scan into a grid of
continuous patch embeddings. (ii) A quantizer maps each (projected) embedding to the
nearest code of a discrete codebook $\mathcal{C}=\{\mathbf{w}_k\}_{k=1}^{K}$,
whose codes and connecting topology are maintained by GNG. (iii) A shallow decoder $\mathrm{Dec}_\psi$ (with parameters $\psi$)
reconstructs, from the quantized codes, the semantic feature field produced by
$T$. The tokenizer encoder is distinct from the stage-2 student, a more compact
backbone trained from scratch (Section~\ref{subsec:mim_models}); because the
tokenizer only emits training targets and is never deployed, we keep its encoder
at full (Base) capacity without affecting the inference cost of the final model.

\paragraph{Encoder and forward quantization}
Let $\mathbf{x}\in\mathbb{R}^{H\times W}$ denote an input B-scan, with $H$ and
$W$ its height and width in pixels. The tokenizer
encoder $\mathrm{Enc}_\phi$ is a Vision Transformer that partitions $\mathbf{x}$ into a regular grid of $N=h\times w$ non-overlapping, square patches of side $P$ pixels, where $h=H/P$ and $w=W/P$ are the patch-grid height and width, and outputs continuous embeddings $\{\mathbf{u}_i\}_{i=1}^{N}$ in $\mathbb{R}^{D}$. A linear projection then maps these embeddings to a quantization space of dimension $C_{\mathrm{qdim}}=256$, yielding latents $\{\mathbf{z}_i\}_{i=1}^{N}$, $\mathbf{z}_i\in\mathbb{R}^{C_{\mathrm{qdim}}}$. Both the projected latents and every code are
$\ell_2$-normalized onto the unit hypersphere (the codebook is renormalized after each GNG update), so all nearest-neighbor and reconstruction computations are carried out in cosine geometry; normalization also removes codebook-magnitude bias, whereby a few high-norm codes can dominate nearest-neighbor assignment \citep{yu2022vector}. Each latent is discretized by assigning it to its nearest code under the cosine distance,
\begin{equation}
\label{eq:winner}
k_i^\star \;=\; \arg\min_{k\in\{1,\dots,K\}} d_{\cos}(\mathbf{z}_i,\mathbf{w}_k),
\quad
d_{\cos}(\mathbf{a},\mathbf{b}) \;=\; 1-\frac{\mathbf{a}^\top\mathbf{b}}{\lVert\mathbf{a}\rVert\,\lVert\mathbf{b}\rVert},
\end{equation}
and is replaced by the corresponding code, $\mathbf{z}_i^{q}=\mathbf{w}_{k_i^\star}$;
here $K=|\mathcal{V}|$ is the \emph{current} codebook size, which grows from
$n_0$ to $K_{\max}$ as the GNG graph introduced below expands
(Section~\ref{subsec:implementation_details}). The index $k_i^\star$ is the
discrete visual token of patch $i$.

\paragraph{Teacher-guided reconstruction}
The decoder $\mathrm{Dec}_\psi$ ingests the quantized grid $\{\tilde{\mathbf{z}}_i^{q}\}$
(the straight-through codes of Eq.~\eqref{eq:ste}) and
predicts, for every patch, the semantic embedding of the frozen teacher rather
than raw pixels; being shallow, it maps the quantization dimension back to the
embedding dimension, $C_{\mathrm{qdim}}\!\to\!D$. Let $\mathbf{t}_i = T(\mathbf{x})_i \in
\mathbb{R}^{D}$ be the teacher embedding of patch $i$ and $\mathbf{o}_i =
\mathrm{Dec}_\psi(\{\tilde{\mathbf{z}}_j^{q}\})_i$ the corresponding decoder output. The teacher
feature field is spatially aligned to the encoder's $N=h\times w$ grid: a
teacher whose native token grid is finer than $N$ is average-pooled to match
(e.g.\ DINOv3's $32\times32$ tokens are pooled to $16\times16$), whereas a
teacher already on the $N$ grid (e.g.\ MIRAGE) is used directly. The tokenizer
is trained to maximize the directional agreement between the two,
\begin{equation}
\label{eq:rec}
\mathcal{L}_{\mathrm{rec}}
\;=\; \frac{1}{N}\sum_{i=1}^{N} d_{\cos}\!\bigl(\mathbf{o}_i,\mathbf{t}_i\bigr),
\end{equation}
using the cosine distance $d_{\cos}(\cdot,\cdot)$ defined in Eq.~\eqref{eq:winner}, so that the discrete codes are forced to retain the semantic content of the
teacher representation space.

\paragraph{Optimization constraint}
The assignment in Eq.~\eqref{eq:winner} is non-differentiable. To propagate
gradients from the reconstruction loss back to the encoder we use the
straight-through estimator (STE; \citealp{bengio2013estimating,vandenoord2017neural}), copying the gradient
of the quantized code directly to the latent,
\begin{equation}
\label{eq:ste}
\tilde{\mathbf{z}}_i^{q} \;=\; \mathbf{z}_i + \mathrm{sg}\!\left[\mathbf{z}_i^{q}-\mathbf{z}_i\right],
\end{equation}
where $\mathrm{sg}[\cdot]$ denotes the stop-gradient operator, so that
$\tilde{\mathbf{z}}_i^{q}$ equals $\mathbf{z}_i^{q}$ on the forward pass but
passes the gradient to the encoder on the backward pass. A commitment
term keeps the encoder outputs close to the selected codes. The complete
tokenizer objective is
\begin{equation}
\label{eq:total}
\mathcal{L}_{\mathrm{tok}}
\;=\; \underbrace{\frac{1}{N}\sum_{i=1}^{N}d_{\cos}\!\bigl(\mathbf{o}_i,\mathbf{t}_i\bigr)}_{\text{teacher reconstruction}}
\;+\;
\beta\,\underbrace{\frac{1}{N}\sum_{i=1}^{N}\bigl\lVert \mathbf{z}_i-\mathrm{sg}\!\left[\mathbf{w}_{k_i^\star}\right]\bigr\rVert_2^{2}}_{\text{commitment}},
\end{equation}
with $\beta$ weighting the commitment term. Crucially, Eq.~\eqref{eq:total}
contains \emph{no} codebook (dictionary) loss term: the codes
$\{\mathbf{w}_k\}$ are never moved by gradient descent. Their values, number,
and connectivity are governed entirely by the GNG procedure described next.

\paragraph{Codebook learning by Growing Neural Gas}
The codebook is treated as a graph $\mathcal{G}=(\mathcal{V},\mathcal{E})$, with
$\mathcal{E}$ the set of edges, in
which each vertex $k$ carries a code vector $\mathbf{w}_k$ and an accumulated
error $E_k$, and edges encode neighborhood relations between codes. The graph
is initialized with a small number $n_0$ of vertices whose code vectors are
random unit-norm vectors and no edges ($\mathcal{E}=\emptyset$). It is grown \emph{concurrently} with encoder
training: at every optimization step the current encoder latents---detached
from the computation graph---are presented to the GNG as input signals, so the
codebook continually tracks the evolving encoder rather than being fitted once
to a frozen encoder. A \emph{signal} here is a single patch latent
$\mathbf{z}_i$; every patch of every image in the batch contributes one
signal, so a batch of $B$ images yields $B\!\times\!N$ signals per step.
Algorithm~\ref{alg:gng} details the full update; all steps are applied per signal except the error decay, which is applied once per training batch; we summarize its three key ideas below.

\begin{algorithm*}[t]
\caption{Growing Neural Gas codebook update}
\label{alg:gng}
\begin{algorithmic}[1]
\Require patch latents $\{\boldsymbol{\xi}\}$ (one per patch), grouped into training batches; graph $\mathcal{G}=(\mathcal{V},\mathcal{E})$ initialized with $n_0$ random unit-norm codes $\{\mathbf{w}_k\}$ and errors $\{E_k\}$
\Statex \textbf{Hyperparameters:} $\varepsilon_b,\varepsilon_n,\alpha,\delta,a_{\max},\lambda,K_{\max}$
\For{each training batch $\mathcal{B}$}
\For{each signal $\boldsymbol{\xi}\in\mathcal{B}$}
    \State $s_1,s_2 \gets$ two nearest units to $\boldsymbol{\xi}$ under $d_{\cos}(\cdot,\cdot)$ \Comment{Eq.~\eqref{eq:winner}}
    \State $E_{s_1} \gets E_{s_1} + d_{\cos}(\boldsymbol{\xi},\mathbf{w}_{s_1})$ \Comment{accumulate error}
    \State $\mathbf{w}_{s_1} \gets \mathbf{w}_{s_1} + \varepsilon_b(\boldsymbol{\xi}-\mathbf{w}_{s_1})$ \Comment{move winner, Eq.~\eqref{eq:move_winner}}
    \For{$n \in \mathcal{N}(s_1)$}
        \State $\mathbf{w}_{n} \gets \mathbf{w}_{n} + \varepsilon_n(\boldsymbol{\xi}-\mathbf{w}_{n})$ \Comment{move neighbors, Eq.~\eqref{eq:move_neighbors}}
    \EndFor
    \State $\mathbf{w}_{s_1} \gets \mathbf{w}_{s_1}/\lVert\mathbf{w}_{s_1}\rVert$;\;\; $\mathbf{w}_{n} \gets \mathbf{w}_{n}/\lVert\mathbf{w}_{n}\rVert\;\;\forall n\in\mathcal{N}(s_1)$ \Comment{$\ell_2$-renormalize moved codes}
    \State create/reset edge $(s_1,s_2)$ to age $0$; increment age of other edges of $s_1$
    \State remove edges with age $> a_{\max}$ and any isolated vertices
    \If{$\lambda$ signals elapsed \textbf{and} $|\mathcal{V}| < K_{\max}$}
        \State $u \gets \arg\max_k E_k$;\;\; $f \gets \arg\max_{n\in\mathcal{N}(u)} E_n$
        \State create new vertex $r$ with $\mathbf{w}_r=\tfrac{1}{2}(\mathbf{w}_u+\mathbf{w}_f)$, then $\mathbf{w}_r \gets \mathbf{w}_r/\lVert\mathbf{w}_r\rVert$ ($\ell_2$-renormalize); rewire $u\!-\!r$, $f\!-\!r$ \Comment{Eq.~\eqref{eq:insert}}
        \State $E_u \gets \alpha E_u$;\;\; $E_f \gets \alpha E_f$;\;\; $E_r \gets E_u$
    \EndIf
\EndFor
\State $E_k \gets \delta\,E_k \quad \forall k$ \Comment{global error decay, applied once per batch after all signals in $\mathcal{B}$, not per signal}
\EndFor
\State \Return updated graph $\mathcal{G}$
\end{algorithmic}
\end{algorithm*}

The winner $s_1$ and runner-up $s_2$ are found with the same nearest-neighbor
computation as in Eq.~\eqref{eq:winner}. \textbf{Winner/neighbor movement:}
rather than updating only the assigned code, as in exponential-moving-average
vector quantization, GNG relocates the winner \emph{together with all of its
topological neighbors} toward the signal,
\begin{align}
\label{eq:move_winner}
\Delta\mathbf{w}_{s_1} &= \varepsilon_b\,(\boldsymbol{\xi}-\mathbf{w}_{s_1}), \\
\label{eq:move_neighbors}
\Delta\mathbf{w}_{n} &= \varepsilon_n\,(\boldsymbol{\xi}-\mathbf{w}_{n}),
\qquad \forall\, n\in\mathcal{N}(s_1),
\end{align}
with adaptation rates $\varepsilon_b\gg\varepsilon_n$ and
$\mathcal{N}(s_1)=\{\,n:(s_1,n)\in\mathcal{E}\,\}$ the graph neighbors of the
winner.
\textbf{Edge creation:} competitive Hebbian learning links codes that jointly
win for the same signal and prunes edges older than $a_{\max}$ (with any
resulting isolated vertices); because all vectors have unit norm,
$\arg\min_k\lVert\boldsymbol{\xi}-\mathbf{w}_k\rVert_2^2=\arg\min_k d_{\cos}(\boldsymbol{\xi},\mathbf{w}_k)$,
so adjacency in $\mathcal{G}$ reflects proximity in the cosine metric used for
tokenization. \textbf{Vertex insertion:} each winner accumulates its
quantization error (Algorithm~\ref{alg:gng}), and every $\lambda$ signals a new
unit is inserted between the highest-error vertex $u$ and its highest-error
neighbor $f$,
\begin{equation}
\label{eq:insert}
\mathbf{w}_{r}=\tfrac{1}{2}\bigl(\mathbf{w}_{u}+\mathbf{w}_{f}\bigr),
\qquad
E_u\leftarrow\alpha E_u,\;\; E_f\leftarrow\alpha E_f,\;\; E_r\leftarrow E_u,
\end{equation}
with $\alpha\in(0,1)$ keeping the same pair from being selected again
immediately; all errors additionally decay by the factor $\delta$ once per
training batch, biasing growth toward persistently poorly modeled regions.
Insertion stops once the budget $K_{\max}=512$ is reached, while movement, edge
updates, and pruning continue.

\subsection{Masked image modeling with GNG graph guidance}\label{subsec:mim_models}
After stage 1 has converged, the tokenizer encoder, the codebook $\mathcal{C}$,
and its graph $\mathcal{G}$ are frozen, and the decoder $\mathrm{Dec}_\psi$ is
discarded. The second stage then pretrains the student by masked image modeling
(MIM): the frozen tokenizer supplies the discrete target tokens of each B-scan
via Eq.~\eqref{eq:winner}, and $\mathcal{G}$ supplies the topological prior for
the soft targets.
We build directly on the BEiT-2 framework \citep{Peng2023BEiTv2} and inherit its
two defining components: patch-wise prediction of VQ-KD codebook indices at
masked positions, and a patch-aggregation branch that concentrates global
information in the class token. Our contribution enters through the training
signal rather than the architecture: instead of regressing each masked patch to
a single ground-truth code under a flat one-hot objective, we replace the
one-hot target with a soft target distribution derived from the GNG adjacency graph $\mathcal{G}=(\mathcal{V},\mathcal{E})$ learned in
stage 1, so that supervision is graded by topological proximity within the code
vocabulary. Every other element of the pipeline follows BEiT-2 (see Fig.~\ref{fig:framework}-B).

\paragraph{Input corruption and masking}
Following BEiT-2 \citep{Peng2023BEiTv2}, each B-scan is augmented
once, producing a single view $\mathbf{x}$ that is shared identically by both
streams: it is passed, unmasked, to the frozen tokenizer to generate the
target token grid, and separately patch-embedded into the same
$N=h\times w=16\times 16=256$ patch grid used by the tokenizer, producing patch embeddings $\{\mathbf{x}_i\}_{i=1}^{N}$ for the backbone. A block-wise mask \citep{Peng2023BEiTv2} draws contiguous rectangular blocks until roughly $40\%$ of the grid (about $102$ of the $256$ patches) is selected, defining the masked index set $\mathcal{M}\subset\{1,\dots,N\}$. Each embedding is then
corrupted by a single learnable mask token $\mathbf{e}_{[\mathrm{M}]}$,
\begin{equation}
\label{eq:mask}
\mathbf{x}_i^{\mathcal{M}} \;=\; (1-m_i)\,\mathbf{x}_i \;+\; m_i\,\mathbf{e}_{[\mathrm{M}]},
\qquad m_i=\mathds{1}[\,i\in\mathcal{M}\,],
\end{equation}
so that only masked positions are overwritten. A learnable class token
$\mathbf{e}_{[\mathrm{CLS}]}$ is prepended, giving the length-$(N{+}1)$ input
sequence $\bigl[\mathbf{e}_{[\mathrm{CLS}]},\,\{\mathbf{x}_i^{\mathcal{M}}\}_{i=1}^{N}\bigr]$
that is fed to the ViT. Position is encoded by a shared relative position bias
added to every attention map, in place of absolute position embeddings.

\paragraph{Prediction pipeline and patch-level objective}
The corrupted sequence is processed by the ViT backbone, whose internal blocks
we treat as a standard black box; let $\mathbf{h}_i^{L}$ denote the final-layer
hidden state of patch $i$. A lightweight prediction head---a single
fully-connected layer $\mathbf{W}$---maps each masked patch state to a
categorical distribution over the $K$ active codes of the frozen codebook,
\begin{equation}
\label{eq:mim_head}
p\bigl(\cdot \mid \mathbf{x}_i^{\mathcal{M}}\bigr)
\;=\;\mathrm{softmax}\!\bigl(\mathbf{W}\,\mathbf{h}_i^{L}\bigr).
\end{equation}
Writing $k_i^\star$ for the code index emitted by the frozen tokenizer at patch
$i$ (Eq.~\eqref{eq:winner}), the standard BEiT-2 objective is the one-hot
cross-entropy evaluated only at masked positions,
\begin{equation}
\label{eq:mim_hard}
\mathcal{L}_{\mathrm{MIM}}
\;=\; -\sum_{i\in\mathcal{M}} \log p\bigl(k_i^\star \mid \mathbf{x}_i^{\mathcal{M}}\bigr).
\end{equation}
Equation~\eqref{eq:mim_hard} is our baseline and the departure point for the
graph-guided target introduced next.

\paragraph{GNG graph-distance soft targets}
A one-hot target treats every incorrect code as equally wrong, discarding the
fact that the stage-1 codebook is not a flat set but a graph: competitive
Hebbian edges connect codes that jointly win for nearby latents
(Section~\ref{subsec:image_tokenizer}), so adjacency in $\mathcal{G}$ reflects
local proximity in the learned quantization space. We exploit this relational
structure by softening the target according to graph geodesic distance. Let
$d_{\mathcal{G}}(a,b)$ be the shortest-path (hop) distance between codes $a$ and
$b$ in $\mathcal{G}$, with $d_{\mathcal{G}}=\infty$ for codes lying in different
connected components. For a ground-truth code $k_i^\star$ we define the soft
target
\begin{equation}
\label{eq:softtarget}
q\bigl(j \mid k_i^\star\bigr)
\;=\;
\frac{\mathds{1}\!\bigl[d_{\mathcal{G}}(k_i^\star,j)\le R\bigr]\,\exp\!\bigl(-\,d_{\mathcal{G}}(k_i^\star,j)/\tau\bigr)}
     {\sum_{k:\,d_{\mathcal{G}}(k_i^\star,k)\le R}\exp\!\bigl(-\,d_{\mathcal{G}}(k_i^\star,k)/\tau\bigr)},
\end{equation}
where the temperature $\tau$ controls how quickly probability mass decays with
graph distance, and the indicator $\mathds{1}[\cdot]$ restricts support to an $R$-hop neighborhood
($R=3$): codes farther than $R$ hops, or in a disconnected component, receive
zero mass by construction, and an isolated code falls back to the one-hot target. Because
$\mathcal{G}$ is fixed after stage 1, the full $K\times K$ geodesic-distance
matrix is precomputed once by breadth-first search and converted to the
soft-target matrix $\mathbf{Q}$, whose row $k_i^\star$ is
Eq.~\eqref{eq:softtarget}; $\mathbf{Q}$ is cached and simply looked up for every
masked patch. The patch-level objective then becomes a soft cross-entropy,
\begin{equation}
\label{eq:mim_soft}
\mathcal{L}^{\mathrm{soft}}_{\mathrm{MIM}}
\;=\; -\sum_{i\in\mathcal{M}}\sum_{j=1}^{K}
q\bigl(j \mid k_i^\star\bigr)\,\log p\bigl(j \mid \mathbf{x}_i^{\mathcal{M}}\bigr),
\end{equation}
which recovers the one-hot loss of Eq.~\eqref{eq:mim_hard} exactly as
$\tau\!\to\!0$. In effect the backbone is penalized less for confusing a
code with its topological neighbors, and the gradient is graded by distance in
the learned code graph, with the aim of transferring the relational structure of
the vocabulary into the representation.

\paragraph{Global aggregation and unified objective}
Optimizing Eq.~\eqref{eq:mim_soft} alone supervises patch-level (local)
predictions and leaves the class token---on which downstream image-level tasks
rely---weakly constrained, the ``global representation bottleneck'' identified
in BEiT-2 \citep{Peng2023BEiTv2}. We adopt its patch-aggregation remedy
unchanged. Patch tokens taken from an intermediate layer $l$ (with $l=9$ for the
ViT-B backbone), $\{\mathbf{h}_i^{l}\}$, are concatenated with the final-layer
class token $\mathbf{h}_{\mathrm{CLS}}^{L}$ into the sequence
\begin{equation}
\label{eq:agg}
\mathbf{S} \;=\; \bigl[\,\mathbf{h}_{\mathrm{CLS}}^{L},\,\mathbf{h}_1^{l},\dots,\mathbf{h}_N^{l}\,\bigr],
\end{equation}
which is passed through a shallow two-layer transformer decoder to yield a
second set of masked-patch predictions
$p^{c}(\cdot \mid \mathbf{x}_i^{\mathcal{M}})$ through the \emph{same} shared
head $\mathbf{W}$ of Eq.~\eqref{eq:mim_head}. These predictions are supervised
by the identical graph-distance soft targets, giving a global soft
cross-entropy $\mathcal{L}^{c,\mathrm{soft}}_{\mathrm{MIM}}$ of the same form as
Eq.~\eqref{eq:mim_soft}. The two heads are trained jointly under the unified
objective
\begin{equation}
\label{eq:mim_total}
\mathcal{L}^{\mathrm{total}}_{\mathrm{MIM}} \;=\; \mathcal{L}^{\mathrm{soft}}_{\mathrm{MIM}}
\;+\; \mathcal{L}^{c,\mathrm{soft}}_{\mathrm{MIM}}.
\end{equation}
Because the intermediate patch tokens in Eq.~\eqref{eq:agg} can influence the
prediction only through the single summarizing class token, the shallow decoder
acts as an information bottleneck that forces a complete, decodable image
representation into $\mathbf{h}_{\mathrm{CLS}}^{L}$. This decoder, like the prediction heads, is discarded after pretraining, and
only the backbone weights are retained for the downstream tasks. The ablation in
Section~\ref{subsec5} isolates the graph term by replacing $\mathbf{Q}$ with the
identity matrix, which recovers the one-hot objective of
Eq.~\eqref{eq:mim_hard} while leaving all other settings unchanged.

\subsection{Implementation details}\label{subsec:implementation_details}
All models were implemented in PyTorch \citep{paszke2019pytorch}, with the
ViT \citep{dosovitskiy2021image} backbones and standard training utilities built on the \texttt{timm}
library (huggingface.co/docs/timm); downstream metrics were computed with
scikit-learn (scikit-learn.org) and, for the Hausdorff distance, with
MONAI \citep{cardoso2022monai}. Across every stage and task, the
input B-scan is resized to $512\times512$ pixels and min-max normalized to
$[0,1]$; with a patch size of $32$ this yields the common $16\times16=256$ patch
grid used by the tokenizer, the backbone, and all decoders; DINOv3 uses a patch size of $16$ ($32\times32$ tokens), whose features are mapped to the common $16\times16$ grid by $2\times2$ average pooling. Both teachers are Base variants (MIRAGE-Base and DINOv3-Base, $D=768$), and unless otherwise noted all encoders are $12$-layer ViTs with $D=768$, $12$ heads, and MLP ratio $4$.

\paragraph{Visual tokenizer (VQ-KD) pretraining}
The decoder $\mathrm{Dec}_\psi$ is a shallow $3$-layer ViT. The tokenizer encoder $\mathrm{Enc}_\phi$ is randomly initialized and trained from scratch, supervised only by the frozen teacher's features, without inheriting the teacher's weights. A separate tokenizer is trained for each teacher on the full pretraining corpus (all $383$ volumes, $\approx\!38{,}300$ B-scans) without a held-out validation split, as in BEiT-2 \citep{Peng2023BEiTv2}; the reconstruction loss converged stably. The codebook budget $K_{\max}=512$ and dimension $C_{\mathrm{qdim}}=256$ were selected jointly on a held-out subset of Kermany's official training data, independent of its test set (Supplementary Table~2). The commitment weight is $\beta=1$, and the GNG graph grows from $n_0=8$ seed vertices to $K_{\max}=512$ codes via a vectorized implementation of the GNG update \citep{Orts-Escolano2015}. The encoder and decoder are optimized with Adam at a base learning rate of $2\times10^{-4}$ for $45$ epochs with a batch size of $128$; full hyperparameters are listed in Supplementary Table~1. Both learned graphs formed a single connected component without isolated nodes, and the DINOv3-derived graph was markedly denser than the MIRAGE-derived one ($5{,}737$ vs.\ $1{,}534$ undirected edges; mean node degree $22.41$ vs.\ $5.99$).

\paragraph{Masked image modeling (MIM) pretraining}
Stage 2 pretrains two student backbones, ViT-S ($D=384$, $6$ heads, otherwise the default architecture) and ViT-B, on the same corpus, with the masking and aggregation branch of Section~\ref{subsec:mim_models} (intermediate layer $l=9$) and the augmentation recipe of the tokenizer (horizontal flip, affine warp, and intensity jitter). The soft targets use $\tau=1.0$ and $R=3$, set once from a sensitivity sweep over $\tau\in\{0.5,1.0,2.0\}$ and $R\in\{2,3,4\}$ on the official Kermany test split with the GNG-ViT-Small-MIRAGE configuration and applied unchanged to every model and dataset (Supplementary Table~3). Optimization uses AdamW with a peak learning rate of $1.5\times10^{-3}$, a batch size of $128$, and $180$ epochs; full hyperparameters are listed in Supplementary Table~1.

\paragraph{Downstream segmentation and classification}
Segmentation was evaluated on Duke iAMD ($4$ classes: background plus the
ILM--RPEDC, RPEDC--BM, and sub-BM regions) and AROI ($8$ classes following \citet{morano2025mirage}: background above the ILM; four
anatomical regions, ILM--IPL/INL, IPL/INL--RPE, RPE--BM, and sub-BM; and three lesion classes, PED, SRF, and IRF (labeled ``cyst'' in the dataset), with one
patient excluded for annotation inconsistency and only the annotated B-scans
retained). For each dataset, a patient-level stratified split with seed $42$ allocated approximately $70\%/10\%/20\%$ of patients to train/validation/test: $17/2/4$ patients for AROI after exclusion and $269/38/76$ patients for Duke iAMD segmentation. The ViT encoder
is kept frozen throughout and feeds a ConvNeXt segmentation decoder
\citep{liu2022convnext, Bachmann2022multimae}, following \citet{morano2025mirage} to ensure a
fair comparison with the MIRAGE baseline; only the decoder was trained, using a
combined cross-entropy and Dice loss. Full segmentation
optimization hyperparameters (optimizer, learning rate, batch size,
augmentation, checkpoint selection, and early stopping) are listed in Supplementary Table~1.

For classification we followed the linear-probing protocol of
\citet{morano2025mirage}: the pretrained encoder is frozen, and only a single linear layer with a softmax
activation is trained on top with a cross-entropy loss and label smoothing of
$0.1$ (full hyperparameters in Supplementary Table~1). Kermany, which is
distributed as individual B-scans rather than volumes, is used in full with its
official author-provided train/validation/test split, the validation subset
serving only for early stopping. For Duke iAMD, a patient-level stratified split with seed $42$ allocated approximately $70\%/10\%/20\%$ of subjects to train/validation/test ($267/39/77$ subjects); for OLIVES, a label-stratified patient-level split (seed $42$) allocated $60/9/18$ patients ($1{,}134/176/284$ volumes). All downstream experiments ran on a single NVIDIA GeForce RTX 4090 GPU (24~GB) with $128$~GB of system RAM.

\subsection{Evaluation metrics}\label{subsec:cls_evaluation}\label{subsec:seg_evaluation}
Following \citet{morano2025mirage}, downstream classification performance was quantified using AUROC (one-vs-rest, weighted by the number of positive samples per class; \citealp{provost2000well,fawcett2006introduction}), average precision (AP, analogously weighted), and balanced accuracy (BAcc, the mean of the per-class recall values; \citealp{mosley2013balanced}). Each classification model was trained with five random seeds, which vary only the linear probe's weight initialization and mini-batch order while the frozen encoder and the data split stay fixed; we report the mean and standard deviation of each metric across the five runs. For the graph-component ablation (Section~\ref{subsec5}), statistical significance between each GNG variant and its matched OneHot baseline of the same backbone size was assessed with a paired, two-tailed Student's $t$-test across the five matched seeds, with differences considered nominally significant at $p<0.05$ before correction. Because this ablation involves 18 simultaneous comparisons (three metrics $\times$ three datasets $\times$ two backbone sizes), we additionally applied a Benjamini--Hochberg false-discovery-rate correction (BH-FDR) across all 18 tests, reporting three significance tiers ($q<0.05$, $q<0.01$, $q<0.001$); BH-FDR was preferred over a family-wise correction because the tests are correlated and the ablation is exploratory. Significance markers in Table~\ref{tab:abl_cls} reflect this correction. Because each reported GNG and OneHot encoder comes from a single pretraining run, these tests quantify probe-level variability conditional on that run and do not estimate variability arising from pretraining stochasticity (e.g.\ the random GNG seed vertices, signal presentation order, and insertion timing). Segmentation accuracy was quantified using the Dice coefficient, intersection-over-union (IoU), and the 95th-percentile Hausdorff distance (HD95), following \citet{morano2025mirage}.

For each dataset, the models were trained to segment all available classes simultaneously: on AROI this comprised both retinal layers and lesions, whereas Duke iAMD contained retinal layer annotations only. Each segmentation model was trained once; the reported standard deviations describe patient-level variability. Dice, IoU, and HD95 were computed for each foreground class on every B-scan and averaged over the B-scans of each patient; HD95 is reported in in-plane pixels. Background was excluded from the reported metrics, and undefined scores were excluded from averaging. For group-level reporting, class scores were averaged within each patient, separately for layers and lesions where applicable, before calculating the reported means and standard deviations across patients. As in \citet{morano2025mirage}, the sub-BM region is treated as a layer class in both datasets and is therefore included in the layer-level metrics.

\section{Results}\label{sec2}
\subsection{Ocular disease classification}\label{subsec3}
After self-supervised pretraining on Duke iAMD, we transferred each MIM encoder
to three downstream OCT classification benchmarks---Kermany, Duke iAMD, and
OLIVES---and report area under the ROC curve (AUROC), average precision (AP),
and balanced accuracy (BAcc) in Table~\ref{tab:cls}. We compare our distilled
models, GNG-ViT (Base and Small), against their respective teachers,
MIRAGE and DINOv3 (both teachers are Base). All encoders were kept frozen; only a single linear layer was trained on top
of the extracted representations (linear probing; \citealp{morano2025mirage}).

The compressed GNG-ViT-Small-MIRAGE, with roughly one-quarter of the teacher's parameters, exceeded MIRAGE in $7$ of $9$ dataset--metric comparisons---all three metrics on both Duke iAMD and OLIVES (e.g.\ OLIVES BAcc $95.22\rightarrow98.06$) and AP on Kermany---and stayed close on Kermany AUROC and BAcc ($98.33$ vs.\ $98.48$; $84.30$ vs.\ $85.32$). The same-size GNG-ViT-Base-MIRAGE exceeded MIRAGE in all $9$, most clearly on four-class Kermany (BAcc $85.32\rightarrow89.98$), so that distillation with the MIRAGE teacher improved on it in $16$ of $18$ comparisons overall.

When distilled from DINOv3, GNG-ViT achieved some of the strongest results on Duke
iAMD and OLIVES. GNG-ViT-Small-DINOv3 obtained the highest balanced accuracy on
Duke iAMD ($93.05$ vs.\ $90.69$ for DINOv3), and GNG-ViT-Base-DINOv3 the highest
AUROC and AP on OLIVES ($99.74$ and $99.82$),
although neither surpassed DINOv3 on Kermany.
Beyond mean performance, distillation also reduced linear-probe variability
conditional on the fixed pretrained encoder: on Duke iAMD, the
balanced-accuracy standard deviation across probe seeds fell from $\pm5.05$
(DINOv3) to $\pm0.61$ (GNG-ViT-Small-DINOv3).

\begin{table*}[!tb]
\centering
\caption{Downstream classification after self-supervised pretraining on
Duke iAMD. Values are mean\,$\pm$\,s.d.\ (\%). Best per column within each
dataset is in bold. GNG-ViT models are distilled from the teacher named in
their suffix (MIRAGE or DINOv3).}
\label{tab:cls}
\begin{adjustbox}{max width=\linewidth}
\begin{tabular}{lccc}
\toprule
Model & AUROC (\%) & AP (\%) & BAcc (\%) \\
\midrule
\multicolumn{4}{l}{\textit{Kermany (4-class: CNV, DME, Drusen, Normal)}}\\
MIRAGE            & 98.48\,$\pm$\,0.05 & 95.91\,$\pm$\,0.14 & 85.32\,$\pm$\,0.39 \\
GNG-ViT-Base-MIRAGE    & 99.44\,$\pm$\,0.04 & 98.56\,$\pm$\,0.09 & \textbf{89.98\,$\pm$\,0.55} \\
GNG-ViT-Small-MIRAGE   & 98.33\,$\pm$\,0.18 & 96.09\,$\pm$\,0.38 & 84.30\,$\pm$\,1.17 \\
DINOv3                 & \textbf{99.69\,$\pm$\,0.08} & \textbf{99.17\,$\pm$\,0.21} & 89.26\,$\pm$\,3.16 \\
GNG-ViT-Base-DINOv3    & 98.35\,$\pm$\,0.17 & 96.04\,$\pm$\,0.42 & 81.57\,$\pm$\,1.11 \\
GNG-ViT-Small-DINOv3   & 99.02\,$\pm$\,0.09 & 97.48\,$\pm$\,0.28 & 82.77\,$\pm$\,2.13 \\
\midrule
\multicolumn{4}{l}{\textit{Duke iAMD (2-class: Control vs.\ iAMD)}}\\
MIRAGE            & 95.76\,$\pm$\,0.80 & 97.03\,$\pm$\,0.97 & 88.08\,$\pm$\,0.97 \\
GNG-ViT-Base-MIRAGE    & 97.09\,$\pm$\,0.26 & 98.65\,$\pm$\,0.13 & 88.33\,$\pm$\,2.46 \\
GNG-ViT-Small-MIRAGE   & \textbf{98.95\,$\pm$\,0.38} & \textbf{99.54\,$\pm$\,0.19} & 90.00\,$\pm$\,2.95 \\
DINOv3                 & 98.41\,$\pm$\,1.17 & 99.32\,$\pm$\,0.50 & 90.69\,$\pm$\,5.05 \\
GNG-ViT-Base-DINOv3    & 97.57\,$\pm$\,0.18 & 98.90\,$\pm$\,0.10 & 90.43\,$\pm$\,1.74 \\
GNG-ViT-Small-DINOv3   & 98.86\,$\pm$\,0.28 & 99.51\,$\pm$\,0.12 & \textbf{93.05\,$\pm$\,0.61} \\
\midrule
\multicolumn{4}{l}{\textit{OLIVES (2-class: DME vs.\ DR)}}\\
MIRAGE            & 98.74\,$\pm$\,0.45 & 99.10\,$\pm$\,0.35 & 95.22\,$\pm$\,1.99 \\
GNG-ViT-Base-MIRAGE    & 99.53\,$\pm$\,0.08 & 99.66\,$\pm$\,0.05 & 96.78\,$\pm$\,0.89 \\
GNG-ViT-Small-MIRAGE   & 99.31\,$\pm$\,0.07 & 99.53\,$\pm$\,0.04 & 98.06\,$\pm$\,0.27 \\
DINOv3                 & 98.85\,$\pm$\,0.29 & 98.88\,$\pm$\,0.30 & 92.79\,$\pm$\,1.11 \\
GNG-ViT-Base-DINOv3    & \textbf{99.74\,$\pm$\,0.14} & \textbf{99.82\,$\pm$\,0.09} & \textbf{98.22\,$\pm$\,0.26} \\
GNG-ViT-Small-DINOv3   & 99.21\,$\pm$\,0.21 & 99.53\,$\pm$\,0.11 & 97.60\,$\pm$\,0.41 \\
\bottomrule
\end{tabular}
\end{adjustbox}
\end{table*}

\subsection{Segmentation of retinal layers and lesions}\label{subsec4}
\begin{figure*}[!htbp]
    \centering
    \includegraphics[width=1.0\textwidth]{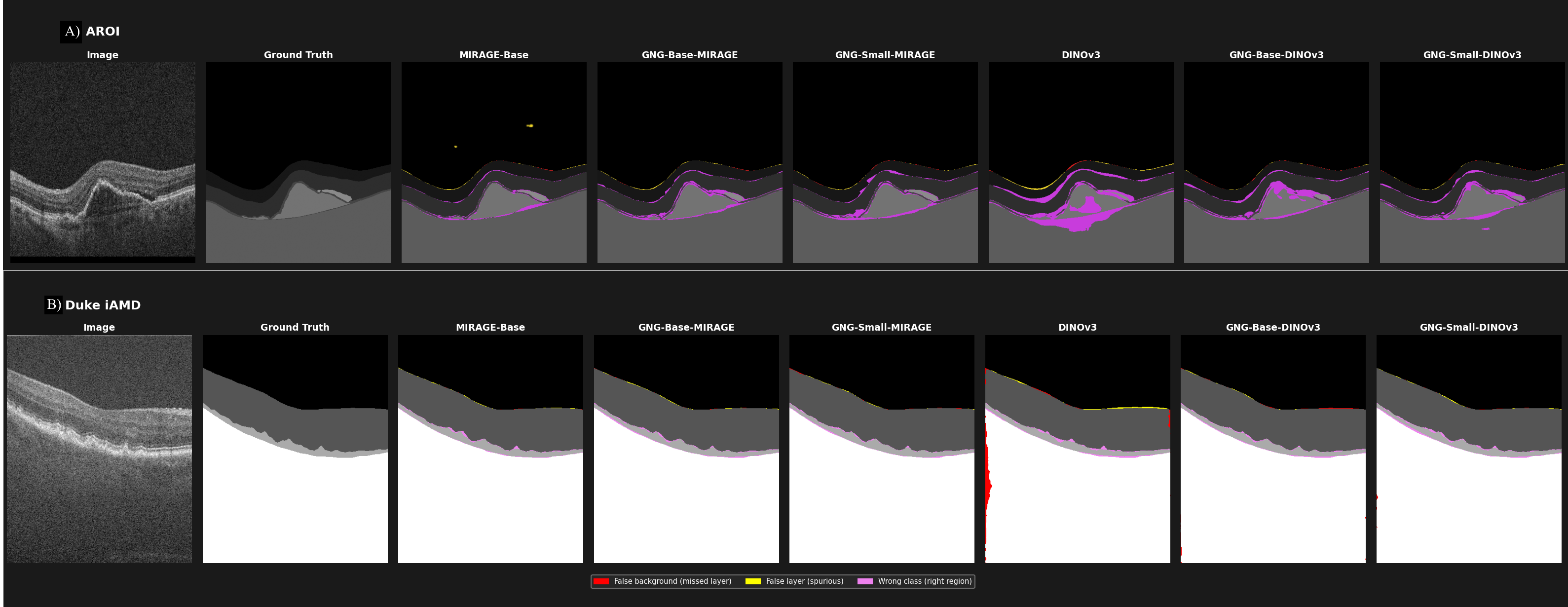}
    \caption{\textbf{Examples of segmentations from different models.} The examples belong to (A) AROI and (B) Duke iAMD. True positives are depicted in the grayscale value of the class; false background pixels (missed layer), in red; false layer pixels (spurious), in yellow; and true but wrongly classified pixels (wrong class, right region), in violet. In these examples, distillation with GNG reduces the segmentation errors of the DINOv3 teacher: on AROI, DINOv3 misclassifies a large fraction of the neurosensory retina as the wrong layer (extensive violet region), an error that is markedly reduced in GNG-Base-DINOv3 and nearly resolved in GNG-Small-DINOv3; on Duke iAMD, DINOv3 exhibits a pronounced false-background strip along the scan edge (red), which is likewise suppressed after distillation. In contrast, MIRAGE and its GNG-distilled variants (GNG-Base-MIRAGE, GNG-Small-MIRAGE) produce visually similar segmentations across both datasets, limited to minor violet boundary slivers, in line with their closely matched quantitative performance.}
    \label{fig:fig_seg_res}
\end{figure*}

We next evaluated the pretrained MIM encoders on two structural OCT segmentation
benchmarks---AROI \citep{Melinscak2021AROI} and Duke iAMD \citep{Farsiu2014AMD}---reporting Dice, intersection over union
(IoU), and the 95th-percentile Hausdorff distance (HD95) in
Table~\ref{tab:seg}. In all experiments the encoder was kept frozen and only
the segmentation decoder was trained, directly assessing the quality of the
pretrained representations \citep{morano2025mirage}. For AROI we separate retinal-layer targets from
focal lesions, as the two differ markedly in size and class balance.
Representative examples for each dataset are shown in
Fig.~\ref{fig:fig_seg_res}.

The benefit of distillation was clearest relative to the DINOv3 teacher.
On AROI retinal layers, GNG-ViT-Small-DINOv3 raised Dice from $83.76$ to
$89.03$ and IoU from $75.83$ to $82.58$, while the same pattern held on
Duke iAMD (Dice $85.34\rightarrow88.34$). Beyond region overlap, distillation
markedly improved boundary localization: DINOv3 produced large HD95 values---$8.02\pm6.17$ on AROI layers and $19.16\pm9.07$ on
Duke iAMD---which GNG-ViT reduced to $1.95\pm0.63$ and $8.40\pm5.01$,
respectively, a roughly two- to four-fold reduction.
A similar reduction was observed for lesion boundaries on AROI, where
HD95 fell from $48.88$ (DINOv3) to $35.74$ (GNG-ViT-Base-DINOv3). 

Distillation from the MIRAGE teacher preserved its already strong layer segmentation and improved lesion delineation. Notably, GNG-ViT-Small-MIRAGE remained within less than one Dice point of the MIRAGE teacher on retinal layers for both datasets (AROI: $-0.49$; Duke iAMD: $-0.64$), nearly reproducing its layer segmentation despite the reduced model size. GNG-ViT-Base-MIRAGE trailed by a slightly larger margin ($-1.30$ and $-1.04$, respectively) with comparable AROI layer boundary error (HD95 $3.18$ vs.\ $2.89$ for MIRAGE).

On the harder AROI lesion targets, GNG-ViT-Base-MIRAGE achieved the best Dice and IoU in Table~\ref{tab:seg} ($51.13$ and $42.16$), while GNG-ViT-Small-MIRAGE also surpassed the MIRAGE teacher ($48.75$ and $40.28$ vs.\ $46.29$ and $38.66$), indicating that the results are consistent with a benefit of distillation on lesion delineation for both model sizes.

    {The qualitative examples in Fig.~\ref{fig:fig_seg_res} support these quantitative comparisons.}

\begin{table*}[!tb]
\centering
\caption{Downstream segmentation after self-supervised pretraining on
Duke iAMD. The encoder was frozen and only the decoder was trained. Values are
mean\,$\pm$\,patient-level s.d.\ (\%; HD95 in pixels), based on a single training run per model. Best per column within each panel is
in bold. $\uparrow$ higher is better; $\downarrow$ lower is better.}
\label{tab:seg}
\begin{adjustbox}{max width=\linewidth}
\begin{tabular}{lccc}
\toprule
Model & Dice (\%)\,$\uparrow$ & IoU (\%)\,$\uparrow$ & HD95\,$\downarrow$ \\
\midrule
\multicolumn{4}{l}{\textit{AROI --- retinal layers}}\\
MIRAGE            & \textbf{90.30\,$\pm$\,1.43} & \textbf{84.40\,$\pm$\,1.82} & 2.89\,$\pm$\,1.52 \\
GNG-ViT-Base-MIRAGE    & 89.00\,$\pm$\,1.80 & 82.90\,$\pm$\,0.83 & 3.18\,$\pm$\,2.42 \\
GNG-ViT-Small-MIRAGE   & 89.81\,$\pm$\,1.54 & 83.76\,$\pm$\,1.89 & \textbf{1.79\,$\pm$\,1.34} \\
DINOv3                 & 83.76\,$\pm$\,1.22 & 75.83\,$\pm$\,1.05 & 8.02\,$\pm$\,6.17 \\
GNG-ViT-Base-DINOv3    & 88.47\,$\pm$\,1.85 & 82.11\,$\pm$\,2.12 & 2.58\,$\pm$\,2.28 \\
GNG-ViT-Small-DINOv3   & 89.03\,$\pm$\,1.19 & 82.58\,$\pm$\,1.27 & 1.95\,$\pm$\,0.63 \\
\midrule
\multicolumn{4}{l}{\textit{AROI --- lesions (PED, SRF, IRF)}}\\
MIRAGE            & 46.29\,$\pm$\,6.71 & 38.66\,$\pm$\,7.27 & 35.44\,$\pm$\,19.57 \\
GNG-ViT-Base-MIRAGE    & \textbf{51.13\,$\pm$\,17.46} & \textbf{42.16\,$\pm$\,17.33} & 36.18\,$\pm$\,15.52 \\
GNG-ViT-Small-MIRAGE   & 48.75\,$\pm$\,18.95 & 40.28\,$\pm$\,17.41 & \textbf{35.08\,$\pm$\,19.78} \\
DINOv3                 & 37.15\,$\pm$\,16.58 & 28.49\,$\pm$\,13.92 & 48.88\,$\pm$\,37.89 \\
GNG-ViT-Base-DINOv3    & 41.19\,$\pm$\,6.84 & 33.10\,$\pm$\,7.64 & 35.74\,$\pm$\,17.78 \\
GNG-ViT-Small-DINOv3   & 42.05\,$\pm$\,22.40 & 34.36\,$\pm$\,21.16 & 37.05\,$\pm$\,17.40 \\
\midrule
\multicolumn{4}{l}{\textit{Duke iAMD --- retinal layers}}\\
MIRAGE            & \textbf{89.51\,$\pm$\,2.86} & \textbf{82.67\,$\pm$\,4.10} & \textbf{7.74\,$\pm$\,4.95} \\
GNG-ViT-Base-MIRAGE    & 88.47\,$\pm$\,2.94 & 80.82\,$\pm$\,4.27 & 8.65\,$\pm$\,5.23 \\
GNG-ViT-Small-MIRAGE   & 88.87\,$\pm$\,2.81 & 81.33\,$\pm$\,4.13 & 8.43\,$\pm$\,5.27 \\
DINOv3                 & 85.34\,$\pm$\,2.86 & 76.60\,$\pm$\,4.06 & 19.16\,$\pm$\,9.07 \\
GNG-ViT-Base-DINOv3    & 87.80\,$\pm$\,3.23 & 79.98\,$\pm$\,4.63 & 9.38\,$\pm$\,5.73 \\
GNG-ViT-Small-DINOv3   & 88.34\,$\pm$\,2.62 & 80.84\,$\pm$\,3.85 & 8.40\,$\pm$\,5.01 \\
\bottomrule
\end{tabular}
\end{adjustbox}
\end{table*}

\subsection{Ablation study}\label{subsec5}
To dissect the sources of the downstream performance reported above, we
conducted two complementary ablations. First, we isolated the contribution of
the teacher alone by evaluating the frozen stage-1 tokenizer encoder as a
feature extractor without any stage-2 MIM pretraining, on both classification
and segmentation. We then isolated the contribution of the graph-based
component by comparing each GNG-ViT model against a matched variant in which
that component is replaced by standard one-hot targets (OneHot), with the MIRAGE teacher and all other
settings held fixed in both ablations. The OneHot comparison isolates the contribution of the soft graph-distance target within a fixed GNG-learned codebook.

\paragraph{Role of the teacher} Before isolating the graph component, we
first asked how much of the downstream performance can be attributed to the
choice of teacher alone, independent of the second-stage MIM pretraining. To
this end we evaluated the frozen stage-1 tokenizer encoder $\mathrm{Enc}_\phi$
directly---i.e.\ without any stage-2 distillation into a student
backbone---as a feature extractor, under the same linear-probing (Kermany)
and frozen-encoder, decoder-only (AROI, ConvNeXt decoder) protocols used
elsewhere (Tables~\ref{tab:abl_teacher_cls} and~\ref{tab:abl_teacher_seg}). Both tokenizers grew from the same $n_0=8$ seed vertices to the same budget $K_{\max}=512$ codes. Both tokenizer encoders were evaluated on the official Kermany test split. On Kermany classification, the tokenizer encoder distilled from DINOv3 outperformed its MIRAGE counterpart across all three metrics (AUROC $76.73\rightarrow90.64$, AP $58.33\rightarrow77.95$, BAcc
$49.28\rightarrow58.06$)—the same ordering observed between the uncompressed teachers in Table~\ref{tab:cls}. On AROI
segmentation, the ranking reversed: the MIRAGE-derived tokenizer encoder
outperformed the DINOv3-derived one on every metric, for both retinal layers
(Dice $85.88$ vs.\ $84.15$, IoU $77.64$ vs.\ $75.61$, HD95 $5.58$ vs.\ $7.72$)
and lesions (Dice $30.66$ vs.\ $26.25$, IoU $23.50$ vs.\ $20.68$, HD95
$58.53$ vs.\ $69.74$). Notably, both
tokenizer encoders lag far behind the corresponding full pipelines: on
Kermany, even the stronger DINOv3-derived tokenizer encoder (BAcc $58.06$)
trails every GNG-ViT student and both full teachers (BAcc $\geq81.57$,
Table~\ref{tab:cls}); on AROI lesions, the MIRAGE-derived tokenizer encoder
(Dice $30.66$) falls well short of the MIRAGE teacher (Dice $46.29$) and the
GNG-ViT-MIRAGE students (Dice $\geq48.75$, Table~\ref{tab:seg}). This
suggests that, while the identity of the teacher already shapes performance
at the tokenizer stage, most of the downstream gains reported above arise
from the second-stage MIM pretraining rather than from the tokenizer encoder
representation alone.

\paragraph{Classification} Adding the graph component significantly improved 10 of
the 18 classification comparisons and significantly worsened 2, with the largest gains on the Base model and the harder
benchmarks (Table~\ref{tab:abl_cls}). On four-class Kermany, GNG-ViT-Base-MIRAGE
improved balanced accuracy from $73.90$ to $89.98$ ($+16.1$ points; paired, two-tailed $t$-test across 5 seeds, $p<0.0001$), a gain that remains significant after BH-FDR correction, alongside
gains in AUROC ($97.58\rightarrow99.44$) and AP ($94.04\rightarrow98.56$), both likewise significant after correction (Table~\ref{tab:abl_cls}). Much of this gain reflects how poorly the Base model trained under one-hot targets: OneHot-Base trailed OneHot-Small on every classification metric and dataset, most markedly in balanced accuracy on Kermany ($73.90$ vs.\ $85.13$) and OLIVES ($92.48$ vs.\ $98.41$). A second pretraining run of OneHot-Base showed the same deficit; this repeat serves only as a qualitative consistency check, and all reported comparisons use one pretraining run per condition. With graph-distance targets, the Base model overtook the Small one on Kermany ($89.98$ vs.\ $84.30$) and narrowed the gap on OLIVES ($96.78$ vs.\ $98.06$); one possible explanation is that the soft targets regularize the higher-capacity student on the small pretraining corpus.
Numerical improvements were seen across all three metrics on Duke iAMD for both model sizes, but they do not all survive multiple-testing correction: AUROC and AP improved significantly for both backbone sizes, whereas the BAcc gains were not significant after correction for either size---including the Base-model BAcc gain ($85.03\rightarrow88.33$), which is nominally significant uncorrected ($p=0.047$) but does not survive BH-FDR correction, and the Small-model BAcc gain (e.g.\ $85.70\rightarrow90.00$) (Table~\ref{tab:abl_cls}). GNG-ViT-Small did not benefit from the graph component on Kermany or OLIVES balanced accuracy, but neither drop was statistically significant (Kermany $p=0.149$; OLIVES $p=0.147$). GNG-ViT-Small's AUROC/AP also dipped significantly vs.\ OneHot-Small on OLIVES (both $p<0.01$), both of which remain significant after BH-FDR correction, yet still exceeded the full MIRAGE teacher despite roughly $4\times$ fewer parameters. This pattern suggests that the graph signal may be less beneficial for the Small backbone in these settings, where the additional supervision may be harder to exploit.

\paragraph{Segmentation} For segmentation the graph component was associated with modest
improvements in region overlap and lower
boundary error (Table~\ref{tab:abl_seg}). On AROI retinal layers,
GNG-ViT-Small-MIRAGE attained the best Dice ($89.81$) and roughly halved the
boundary error relative to its OneHot counterpart (HD95 $3.61\rightarrow1.79$).
On Duke iAMD layers, GNG variants were slightly better than OneHot on Dice, IoU,
and HD95 for both sizes. Lesion-level metrics on AROI showed high patient-level variability and showed no consistent ordering between the two
variants.

\paragraph{Codebook capacity and soft-target hyperparameters} For the codebook-capacity sweep (Supplementary Table~2), we used a held-out subset of 968 Kermany B-scans (242 per class), sampled without replacement from the official training partition with class stratification (seed $42$). These B-scans were excluded from linear-probe training, and the official test set was not used for this selection. Performance varied substantially across the grid (AUROC $70.26$--$79.93$, AP $43.98$--$60.44$, BAcc $37.42$--$49.77$), and 512 codes with 256-dimensional vectors performed best on every metric and were adopted as our default. We separately examined the sensitivity of the GNG soft-target hyperparameters, temperature $\tau\in\{0.5,1.0,2.0\}$ and hop radius $R\in\{2,3,4\}$, on the official Kermany test split (Supplementary Table~3). Sensitivity depended on the metric: AUROC ranged from $97.78$ to $98.34$, AP from $94.33$ to $96.16$, and BAcc from $79.15$ to $84.30$. At $\tau=1$, $R=2$ and $R=3$ gave nearly identical BAcc ($84.25$ vs.\ $84.30$); we adopted $\tau=1.0$, $R=3$.

\begin{table*}[!htbp]
\centering
\caption{Classification performance of the frozen
stage-1 tokenizer encoder alone (no stage-2 MIM pretraining; $512$ codes), evaluated by
linear probing on the official Kermany test split. Values are mean\,$\pm$\,s.d.\ (\%) over five seeds. Best per
column is in bold.}
\label{tab:abl_teacher_cls}
\footnotesize
\begin{adjustbox}{max width=\linewidth}
\begin{tabular}{lccc}
\toprule
Model & AUROC (\%) & AP (\%) & BAcc (\%) \\
\midrule
Tokenizer encoder-MIRAGE  & 76.73\,$\pm$\,0.85 & 58.33\,$\pm$\,1.19 & 49.28\,$\pm$\,0.74 \\
Tokenizer encoder-DINOv3  & \textbf{90.64\,$\pm$\,0.50} & \textbf{77.95\,$\pm$\,0.84} & \textbf{58.06\,$\pm$\,3.06} \\
\bottomrule
\end{tabular}
\end{adjustbox}
\end{table*}

\begin{table*}[!htbp]
\centering
\caption{Segmentation performance of the frozen
stage-1 tokenizer encoder alone (no stage-2 MIM pretraining) on AROI, with a
ConvNeXt decoder trained on top of the frozen encoder (same tokenizers as
Table~\ref{tab:abl_teacher_cls}). Values are mean\,$\pm$\,patient-level s.d.\
(\%; HD95 in pixels), based on a single training run per model. Best per column within each panel is in bold.
$\uparrow$ higher is better; $\downarrow$ lower is better.}
\label{tab:abl_teacher_seg}
\footnotesize
\begin{adjustbox}{max width=\linewidth}
\begin{tabular}{lccc}
\toprule
Model & Dice (\%)\,$\uparrow$ & IoU (\%)\,$\uparrow$ & HD95\,$\downarrow$ \\
\midrule
\multicolumn{4}{l}{\textit{AROI --- retinal layers}}\\
Tokenizer encoder-MIRAGE  & \textbf{85.88\,$\pm$\,1.19} & \textbf{77.64\,$\pm$\,1.44} & \textbf{5.58\,$\pm$\,2.47} \\
Tokenizer encoder-DINOv3  & 84.15\,$\pm$\,1.53 & 75.61\,$\pm$\,2.33 & 7.72\,$\pm$\,2.65 \\
\midrule
\multicolumn{4}{l}{\textit{AROI --- lesions (PED, SRF, IRF)}}\\
Tokenizer encoder-MIRAGE  & \textbf{30.66\,$\pm$\,7.42} & \textbf{23.50\,$\pm$\,7.06} & \textbf{58.53\,$\pm$\,26.50} \\
Tokenizer encoder-DINOv3  & 26.25\,$\pm$\,9.98 & 20.68\,$\pm$\,9.18 & 69.74\,$\pm$\,30.48 \\
\bottomrule
\end{tabular}
\end{adjustbox}
\end{table*}

\begin{table*}[!htbp]
\centering
\caption{Ablation of the graph component on downstream classification
(MIRAGE teacher; linear probing). Values are mean\,$\pm$\,s.d.\ (\%). Within each matched GNG/OneHot pair, bold indicates the significantly better result, based on a Benjamini--Hochberg false-discovery-rate correction (BH-FDR) applied jointly across all 18 tests in this table (paired, two-tailed Student's $t$-test across 5 seeds); $^{*}$, $^{**}$, and $^{***}$ denote, respectively, $0.01\leq q<0.05$, $0.001\leq q<0.01$, and $q<0.001$ for that test, and neither value is bold nor starred when $q\geq0.05$.}
\label{tab:abl_cls}
\begin{adjustbox}{max width=\linewidth}
\begin{tabular}{lccc}
\toprule
Model & AUROC (\%) & AP (\%) & BAcc (\%) \\
\midrule
\multicolumn{4}{l}{\textit{Kermany (4-class)}}\\
OneHot-ViT-Base-MIRAGE  & 97.58\,$\pm$\,0.17 & 94.04\,$\pm$\,0.41 & 73.90\,$\pm$\,0.94 \\
OneHot-ViT-Small-MIRAGE & 98.12\,$\pm$\,0.10 & 96.10\,$\pm$\,0.20 & 85.13\,$\pm$\,0.90 \\
GNG-ViT-Base-MIRAGE    & \textbf{99.44\,$\pm$\,0.04}$^{***}$ & \textbf{98.56\,$\pm$\,0.09}$^{***}$ & \textbf{89.98\,$\pm$\,0.55}$^{***}$ \\
GNG-ViT-Small-MIRAGE   & 98.33\,$\pm$\,0.18 & 96.09\,$\pm$\,0.38 & 84.30\,$\pm$\,1.17 \\
\midrule
\multicolumn{4}{l}{\textit{Duke iAMD (2-class)}}\\
OneHot-ViT-Base-MIRAGE  & 95.28\,$\pm$\,0.86 & 96.83\,$\pm$\,0.51 & 85.03\,$\pm$\,2.75 \\
OneHot-ViT-Small-MIRAGE & 96.15\,$\pm$\,0.40 & 97.19\,$\pm$\,0.20 & 85.70\,$\pm$\,2.20 \\
GNG-ViT-Base-MIRAGE    & \textbf{97.09\,$\pm$\,0.26}$^{*}$ & \textbf{98.65\,$\pm$\,0.13}$^{**}$ & 88.33\,$\pm$\,2.46 \\
GNG-ViT-Small-MIRAGE   & \textbf{98.95\,$\pm$\,0.38}$^{**}$ & \textbf{99.54\,$\pm$\,0.19}$^{***}$ & 90.00\,$\pm$\,2.95 \\
\midrule
\multicolumn{4}{l}{\textit{OLIVES (2-class)}}\\
OneHot-ViT-Base-MIRAGE  & 98.93\,$\pm$\,0.24 & 99.22\,$\pm$\,0.18 & 92.48\,$\pm$\,1.68 \\
OneHot-ViT-Small-MIRAGE & \textbf{99.66\,$\pm$\,0.06}$^{**}$ & \textbf{99.76\,$\pm$\,0.03}$^{**}$ & 98.41\,$\pm$\,0.33 \\
GNG-ViT-Base-MIRAGE    & \textbf{99.53\,$\pm$\,0.08}$^{**}$ & \textbf{99.66\,$\pm$\,0.05}$^{**}$ & \textbf{96.78\,$\pm$\,0.89}$^{*}$ \\
GNG-ViT-Small-MIRAGE   & 99.31\,$\pm$\,0.07 & 99.53\,$\pm$\,0.04 & 98.06\,$\pm$\,0.27 \\
\bottomrule
\end{tabular}
\end{adjustbox}
\end{table*}

\begin{table*}[!htbp]
\centering
\caption{Ablation of the graph component on downstream segmentation
(MIRAGE teacher; frozen encoder, decoder-only training). Values are
mean\,$\pm$\,patient-level s.d.\ (\%; HD95 in pixels), based on a single training run per model. Best per column within each panel is
in bold. $\uparrow$ higher is better; $\downarrow$ lower is better.}
\label{tab:abl_seg}
\begin{adjustbox}{max width=\linewidth}
\begin{tabular}{lccc}
\toprule
Model & Dice (\%)\,$\uparrow$ & IoU (\%)\,$\uparrow$ & HD95\,$\downarrow$ \\
\midrule
\multicolumn{4}{l}{\textit{AROI --- retinal layers}}\\
OneHot-ViT-Base-MIRAGE  & 89.00\,$\pm$\,1.68 & 82.75\,$\pm$\,2.09 & 3.61\,$\pm$\,4.29 \\
OneHot-ViT-Small-MIRAGE & 89.01\,$\pm$\,1.55 & 82.81\,$\pm$\,1.72 & 3.61\,$\pm$\,4.11 \\
GNG-ViT-Base-MIRAGE    & 89.00\,$\pm$\,1.80 & 82.90\,$\pm$\,0.83 & 3.18\,$\pm$\,2.42 \\
GNG-ViT-Small-MIRAGE   & \textbf{89.81\,$\pm$\,1.54} & \textbf{83.76\,$\pm$\,1.89} & \textbf{1.79\,$\pm$\,1.34} \\
\midrule
\multicolumn{4}{l}{\textit{AROI --- lesions (PED, SRF, IRF)}}\\
OneHot-ViT-Base-MIRAGE  & 47.82\,$\pm$\,15.58 & 38.76\,$\pm$\,14.21 & 37.74\,$\pm$\,15.67 \\
OneHot-ViT-Small-MIRAGE & 50.96\,$\pm$\,17.94 & \textbf{42.29\,$\pm$\,17.66} & \textbf{29.94\,$\pm$\,13.49} \\
GNG-ViT-Base-MIRAGE    & \textbf{51.13\,$\pm$\,17.46} & 42.16\,$\pm$\,17.33 & 36.18\,$\pm$\,15.52 \\
GNG-ViT-Small-MIRAGE   & 48.75\,$\pm$\,18.95 & 40.28\,$\pm$\,17.41 & 35.08\,$\pm$\,19.78 \\
\midrule
\multicolumn{4}{l}{\textit{Duke iAMD --- retinal layers}}\\
OneHot-ViT-Base-MIRAGE  & 88.17\,$\pm$\,3.10 & 80.61\,$\pm$\,4.47 & 8.89\,$\pm$\,5.75 \\
OneHot-ViT-Small-MIRAGE & 88.16\,$\pm$\,2.97 & 80.65\,$\pm$\,4.35 & 8.99\,$\pm$\,5.14 \\
GNG-ViT-Base-MIRAGE    & 88.47\,$\pm$\,2.94 & 80.82\,$\pm$\,4.27 & 8.65\,$\pm$\,5.23 \\
GNG-ViT-Small-MIRAGE   & \textbf{88.87\,$\pm$\,2.81} & \textbf{81.33\,$\pm$\,4.13} & \textbf{8.43\,$\pm$\,5.27} \\
\bottomrule
\end{tabular}
\end{adjustbox}
\end{table*}

\subsection{Inference cost}\label{subsec:cost}
Table~\ref{tab:cost} reports inference cost at the common $512\times512$ input. The ViT-Small students require $6.37$ encoder GFLOPs per image, $3.6\times$ fewer than MIRAGE-Base ($23.25$) and $16.9\times$ fewer than DINOv3-Base ($107.52$), whereas the ViT-Base students cost about as much as MIRAGE-Base ($23.65$). The ViT-Base students are thus not compressed ($87.43$M vs.\ $86.04$M parameters) and serve as a same-size control for studying the training objective; compression claims refer to the ViT-Small students. For classification, the ViT-Small students reduced latency from $2.98$\,ms (MIRAGE) and $9.50$\,ms (DINOv3) to $1.66$\,ms and peak memory from $347.0$ and $377.7$\,MiB to $104.3$\,MiB; for segmentation, the reduction was smaller ($5.59$--$5.61$ vs.\ $6.98$ and $13.53$\,ms; $1162.1$ vs.\ $1403.2$ and $1406.3$\,MiB) because the shared decoder dominates the cost.

\begin{table*}[!htbp]
\centering
\caption{Inference cost at $512\times512$ input. Parameters and GFLOPs count the encoder only (one multiply--add = one FLOP). Latency (median of 300 forward passes) and peak memory were measured on one RTX 4090 at batch size 1 in FP32 (TF32 and cuDNN disabled) and include the classification head or the ConvNeXt segmentation decoder; they are setup-specific, not optimized deployment timings. Students share the cost of their backbone regardless of teacher or pretraining target. $^{\dagger}$Both teachers are ViT-Base ($D=768$); DINOv3 uses $16\times16$-pixel patches and thus processes $1{,}024$ tokens per $512\times512$ image, versus $256$ for the patch-32 encoders, which accounts for its higher GFLOPs and latency.}
\label{tab:cost}
\begin{adjustbox}{max width=\linewidth}
\begin{tabular}{lcccccc}
\toprule
 & & & \multicolumn{2}{c}{Classification} & \multicolumn{2}{c}{Segmentation} \\
\cmidrule(lr){4-5}\cmidrule(lr){6-7}
Model & Params (M) & GFLOPs & Latency (ms) & Memory (MiB) & Latency (ms) & Memory (MiB) \\
\midrule
MIRAGE-Base (teacher) & 86.04 & 23.25 & 2.98 & 347.0 & 6.98 & 1403.2 \\
DINOv3-Base (teacher) & 85.67 & 107.52$^{\dagger}$ & 9.50 & 377.7 & 13.53 & 1406.3 \\
\midrule
ViT-Base students & 87.43 & 23.65 & 3.02 & 358.8 & 6.99--7.01 & 1410.4 \\
ViT-Small students & 22.48 & 6.37 & 1.66 & 104.3 & 5.59--5.61 & 1162.1 \\
\bottomrule
\end{tabular}
\end{adjustbox}
\end{table*}

\FloatBarrier
\section{Discussion and conclusion}\label{sec3}

We set out to determine how much of a continuous vision foundation
model's downstream utility for retinal OCT survives compression into a substantially smaller network.
The proposed topology-aware discrete distillation framework reduces the teacher to a roughly fourfold smaller student. Across three classification and two segmentation benchmarks spanning multiple datasets, imaging devices, and clinical tasks, the student achieved performance close to, and in some settings above, its teacher, although the magnitude and direction of the difference depended on the teacher, task, metric, and student capacity. Three observations stand out. First, most of the teacher's downstream utility survived compression: the ViT-Small MIRAGE student stayed within one Dice point of its teacher on the retinal layers of both segmentation datasets and exceeded it on every metric on the external OLIVES benchmark (balanced accuracy $98.06$ vs.\ $95.22$; Tables~\ref{tab:cls} and~\ref{tab:seg}). Second, the framework delivers these results with two teachers of different training domains, both ViT-Base, consistent with the improvements arising from the distillation procedure rather than from any single backbone, and indicating that the added value of retina-specific pretraining is task-dependent. Third, the graph-distance soft cross-entropy target improved several outcomes, particularly for the Base model, and lowered retinal-layer boundary error, but its effect was task- and capacity-dependent and was neutral or adverse in some Small-model settings (Tables~\ref{tab:abl_cls} and~\ref{tab:abl_seg}). Within our pipeline, the role of GNG is to supply the adjacency graph that defines the graph-distance targets.

Beyond compression, examining how the framework behaves across teachers of different training domains reveals its most consequential property. Two observations recur. First, the choice of teacher shapes downstream performance in a task-dependent way. At the teacher level, the natural-image DINOv3 teacher outperforms the retina-specific MIRAGE teacher on every metric on Kermany and Duke iAMD, whereas on OLIVES the two are close on AUROC and AP and MIRAGE leads on balanced accuracy ($95.22$ vs.\ $92.79$); MIRAGE leads on every segmentation benchmark (Tables~\ref{tab:cls} and~\ref{tab:seg}); the same split is already present at the tokenizer stage (Tables~\ref{tab:abl_teacher_cls} and~\ref{tab:abl_teacher_seg}). This ordering broadly carries through to the distilled students, though not uniformly: on Kermany balanced accuracy, for example, the MIRAGE-distilled students lead (Table~\ref{tab:cls}). These tendencies are consistent with the two pretraining objectives: DINOv3's self-distillation tends to produce globally invariant, linearly separable features well suited to image-level classification \citep{park2023what}, while MIRAGE's retina-specific pretraining confers an advantage on fine, anatomically localized structure \citep{morano2025mirage}. That a general-domain model should nonetheless lead or match a retina-specific one on ocular classification is in line with recent evidence that natural-image foundation models can surpass domain-specific retinal models on detection tasks driven by clear visual patterns---precisely the regime of Kermany, Duke iAMD, and OLIVES \citep{Hou2025Can}. Second, the more striking observation concerns the out-of-domain teacher. When DINOv3---a model not pretrained specifically on retinal data---is passed through our pipeline, the resulting student does not merely approximate its teacher on retinal-layer segmentation; it substantially outperforms it. On retinal layers, region overlap improves on both datasets, boundary error falls roughly two- to fourfold, and the teacher's failure modes---large mislabeled retinal regions and edge artifacts---are progressively suppressed after distillation (Table~\ref{tab:seg}, Fig.~\ref{fig:fig_seg_res}). Crucially, this gain is not attributable to the discrete bottleneck alone. The stage-1 tokenizer encoder derived from DINOv3, evaluated without any second-stage pretraining, reaches only $84.15$ Dice on AROI layers---essentially the raw teacher's level (Table~\ref{tab:abl_teacher_seg}). These results suggest that the gain arises largely in the second-stage MIM pretraining, which lifts performance well above the tokenizer-only level (Table~\ref{tab:seg}). One possible explanation is that quantizing the teacher's continuous features against a codebook grown on retinal data may orient the targets toward retinal structure, while graph-guided MIM trains a fresh backbone to predict those codes. These results are consistent with the framework acting not simply as a smaller MIRAGE but as a pipeline that can convert a generic visual backbone into a competitive retinal one. It also bears on how much is gained by pretraining a retina-specific foundation model: the DINOv3-distilled student reaches $89.03$ Dice on AROI retinal layers against $90.30$ for the MIRAGE teacher (Table~\ref{tab:seg}), so a general-domain backbone passed through this pipeline approaches, without matching, a model pretrained on large retinal corpora. The advantage of retina-specific pretraining is therefore task-dependent: it is small or absent for image-level classification, but persists for dense prediction, most clearly for focal lesions (Table~\ref{tab:seg}).

The results are broadly consistent across student sizes, though not perfectly so. On Duke iAMD, the Small students match or exceed both their Base counterparts and the continuous teachers on AUROC and AP, and on OLIVES all four students exceed their teachers on every metric, with the Base students ahead on AUROC and AP, whereas on the larger Kermany benchmark GNG-ViT-Base-MIRAGE is the strongest distilled variant and the gap to the teacher is larger for the DINOv3-distilled students on balanced accuracy (Table~\ref{tab:cls}). The graph target likewise helped most for the Base model and on the harder benchmarks, and lowered retinal-layer boundary error in all four segmentation comparisons (Tables~\ref{tab:abl_cls} and~\ref{tab:abl_seg}).

At inference, only the frozen student encoder and a lightweight task head are required, and the ViT-Small student has roughly one quarter of the teacher's parameters ($22$M vs.\ $86$M) and needs at least $3.6\times$ fewer encoder FLOPs (Table~\ref{tab:cost}). Because latency and peak memory were measured only on a desktop GPU, these results indicate potential suitability for deployment but do not establish deployment readiness. A second, complementary advantage is data efficiency: the student's two-stage pretraining used only a single small public cohort of $383$ volumes, yet it yields a backbone competitive with a teacher trained on far larger corpora. This lowers the barrier for smaller centers, which could in principle obtain a strong task-specific backbone without assembling the large, multi-source datasets that foundation-model pretraining normally requires.

Two aspects of the evaluation bear on clinical relevance. First, the benchmark tasks mirror distinctions that OCT is used to make in practice, from referral urgency, disease detection, and disease differentiation to the layer and fluid measurements that guide anti-VEGF management \citep{kermany2018identifying, flaxel2020amdppp}. Second, three of the four benchmark datasets are external to the pretraining cohort: Kermany, OLIVES, and AROI were not used for the two-stage self-supervised student pretraining on Duke iAMD and were acquired at independent centers on two other devices (Spectralis and Cirrus versus Bioptigen), so the transfer reported on these three datasets is cross-site and cross-device, whereas Duke iAMD is a within-cohort, transductive evaluation (Section~\ref{subsec:benchmark_datasets}). Kermany was, however, used for codebook-capacity and soft-target hyperparameter selection, so OLIVES and AROI are the only benchmarks untouched by both pretraining and model selection.

Our study has five main limitations. First, the framework is evaluated on a single imaging modality, the OCT B-scan; we do not test whether the same topology-aware discretization transfers to other retinal modalities such as fundus photography or OCT angiography. Second, all of our evaluation uses retrospective, publicly available benchmarks and therefore assesses the representation offline; it is no substitute for the prospective clinical validation that any translation into practice would require. We also did not analyze performance across demographic strata such as age, sex, or ancestry: pretraining draws on four US sites, a single device, and a single dominant disease (intermediate AMD), and a distilled student may also inherit biases of its teacher. Third, on Kermany, the soft-target hyperparameters $\tau$ and $R$ were selected on the official test split and applied unchanged to every model, including the Base configuration behind the $+16.1$-point balanced-accuracy gain; the Kermany results are therefore not an independent confirmation (Section~\ref{subsec5}), and our generalization claims rest on OLIVES and AROI. The codebook-capacity sweep, repeated on a held-out selection set drawn from the training partition, reselected the same configuration, and no other test set was used for any selection. Fourth, each segmentation model was trained once, following the single-run protocol of \citet{morano2025mirage}, so the reported segmentation variability is patient-level rather than run-to-run; the AROI lesion metrics in particular carry high variance given their sparse targets and four-patient test set. Fifth, the ablations isolate the graph-distance target against a matched one-hot control, but we did not compare against conventional VQ tokenizers within the full two-stage pipeline; the ablations also include no alternative graph topology (e.g.\ random or $k$-nearest-neighbor) and no stage-wise separation of GNG quantization, graph guidance, and additional retinal MIM training (e.g.\ via a teacher-free student); teacher-agnosticism, moreover, was tested only with two ViT-Base teachers with $D{=}768$, so other architectures and embedding dimensions remain untested.

Several directions follow from this work. Beyond the missing controls listed above, pretraining could be scaled to multi-device, multi-disease, and multi-institution OCT data, and larger teachers and code budgets could be tested, the latter against the density of the adjacency graph that supplies the soft targets. Relating the learned codes to annotated anatomical and pathological structure could reveal whether the discrete vocabulary captures reusable primitives of retinal OCT. In conclusion, topology-aware discrete distillation compresses a vision foundation model, whether retina-specific or general-domain, into a roughly fourfold smaller retinal student that performs close to, and in some settings above, its teacher, and suggests that the added value of retina-specific pretraining is task-dependent. Beyond the OCT B-scan, the approach could serve as a general recipe for compressing medical vision foundation models.

\section*{CRediT authorship contribution statement}
Ali Yavari: Conceptualization, Formal analysis, Investigation, Methodology, Software, Visualization, Writing -- original draft, Writing -- review \& editing. Tilman Schmoll: Project administration, Supervision, Writing -- review \& editing. Rainer A. Leitgeb: Formal analysis, Methodology, Supervision, Visualization, Writing -- review \& editing. Wolfgang Drexler: Supervision, Writing -- review \& editing. Thomas Schlegl: Conceptualization, Formal analysis, Methodology, Visualization, Writing -- review \& editing.

\section*{Declaration of generative AI and AI-assisted technologies in the manuscript preparation process}
During the preparation of this work, the authors used Claude (Anthropic) only to improve the language and readability of the manuscript; it was not used to generate scientific content or references. All changes were carefully reviewed and revised by the authors, who take full responsibility for the content of the published article.

\section*{Declaration of competing interest}
Tilman Schmoll is an employee of Carl Zeiss Meditec AG (Oberkochen, Germany). Wolfgang Drexler and Rainer A. Leitgeb are consultants for Carl Zeiss Meditec AG and have received financial compensation. All other authors declare no conflicts of interest.

\section*{Data availability}
The pretraining data and one of the downstream benchmarks in this study come from Duke iAMD, a publicly available dataset. The remaining downstream benchmarks---Kermany, OLIVES, and AROI---are also publicly available. The links to the datasets are as follows: Duke iAMD (\url{https://people.duke.edu/~sf59/RPEDC_Ophth_2013_dataset.htm}); Kermany (\url{https://data.mendeley.com/datasets/rscbjbr9sj/2}); OLIVES (\url{https://zenodo.org/records/7105232}); AROI (\url{https://ipg.fer.hr/ipg/resources/oct_image_database}). No in-house or private data were used in this study.

\section*{Code availability}
The code is currently maintained in a private repository and will be made publicly available under a CC BY-NC 4.0 license upon acceptance of this manuscript.

\section*{\texorpdfstring{Funding}{Funding}}
This work was supported by Carl Zeiss Meditec AG. Tilman Schmoll is employed by Carl Zeiss Meditec AG, while Wolfgang Drexler and Rainer A. Leitgeb serve as consultants for Carl Zeiss Meditec AG; their contributions are detailed in the CRediT authorship contribution statement. Beyond these contributions, Carl Zeiss Meditec AG had no involvement in the study design, data collection, data analysis or interpretation, manuscript preparation, or the decision to submit the article for publication.

\bibliographystyle{elsarticle-harv}
\bibliography{cmig_bibliography}
    
\end{document}